\documentclass{article}
\usepackage[preprint]{neurips_2026}
\usepackage[utf8]{inputenc} 
\usepackage[T1]{fontenc}    
\usepackage{hyperref}       
\usepackage{url}            
\usepackage{booktabs}       
\usepackage{amsfonts}       
\usepackage{nicefrac}       
\usepackage{microtype}      
\usepackage{xcolor}        
\usepackage{amsmath}
\usepackage{algorithm}
\usepackage{algorithmic}
\usepackage{multirow}
\usepackage{amssymb}
\usepackage{graphicx}
\usepackage{tikz}
\usepackage{xcolor}
\usepackage{bm}

\usetikzlibrary{shapes.geometric, arrows.meta, positioning, fit, backgrounds, calc, decorations.pathreplacing, decorations.markings}

\newcommand{\xvec}{\mathbf{x}}
\newcommand{\hvec}{\mathbf{h}}
\newcommand{\cvec}{\mathbf{c}}
\newcommand{\yvec}{\mathbf{y}}

\newcommand{\cgm}{g}
\newcommand{\iob}{\mathrm{IOB}}
\newcommand{\Ts}{T_s}

\title{PhysioSeq2Seq: A Hybrid Physiological Digital Twin and Sequence-to-Sequence LSTM for Long-Horizon Glucose Forecasting in Type 1 Diabetes}

\author{
\textbf{Phat Tran}$^{1}$ \qquad
\textbf{Neville Mehta}$^{2}$ \qquad
\textbf{Clara Mosquera-Lopez}$^{2}$ \\
\textbf{Robert H. Dodier}$^{2}$ \qquad
\textbf{Lizhong Chen}$^{1}$ \qquad
\textbf{Peter G. Jacobs}$^{1,2}$ \\
$^{1}$Oregon State University \\
$^{2}$Oregon Health \& Science University \\
}

\begin{document}

\maketitle

\begin{abstract}
  Accurate long-horizon glucose forecasting is critical for automated insulin delivery systems, which help people with type~1 diabetes (T1D) manage their glucose and avoid dangerous hypoglycemia. However, standard recursive long short-term memory (LSTM) networks suffer from systematic negative bias at longer horizons due to error compounding, while purely mechanistic ordinary differential equation (ODE) models fail to generalize across individuals when parameterized at the population level. We propose PhysioSeq2Seq, a hybrid architecture that combines patient-specific physiological modeling with a sequence-to-sequence (Seq2Seq) LSTM. For each glucose segment, twin matching searches a population of 300 parameterized digital twins to identify the best-fitting physiological match from a 3-hour continuous glucose monitoring (CGM) history. The 10 internal ODE state variables of the matched twin are injected as exogenous covariates into both the encoder and decoder of the Seq2Seq LSTM. This simultaneous 48-step prediction strategy eliminates recursive error compounding, while the ODE features provide a physics-grounded constraint that bounds long-horizon drift within physiologically plausible ranges. PhysioSeq2Seq was trained on CGM and insulin data from 348 participants in the Type 1 Diabetes Exercise Initiative (T1DEXI) dataset and evaluated on 74 held-out participants. At the 240-minute horizon, PhysioSeq2Seq achieves a MAE of 39.28~mg/dL and a ME of $-$10.62~mg/dL, reducing bias by 13.89~mg/dL over the recursive LSTM and MAE by 28.62~mg/dL over the ODE-based digital twin. These results show that eliminating architectural feedback and injecting patient-matched physiological states is an effective and clinically meaningful strategy for long-horizon glucose forecasting in T1D.
\end{abstract}


\section{Introduction}
\label{sec:intro}
Type 1 diabetes (T1D) is a chronic autoimmune condition affecting approximately 9.5 million people worldwide~\citep{Ogle2025}, in which autoimmune destruction of pancreatic $\beta$-cells results in absolute insulin deficiency, requiring patients to self-manage their glucose or use automated insulin delivery (AID) systems to manage blood glucose concentrations continuously~\citep{Atkinson2014}. The emergence of continuous glucose monitors (CGMs) and AID systems has transformed T1D management, but their safety and efficacy hinge critically on the ability to forecast blood glucose trajectories over clinically meaningful multi-hour horizons~\citep{Cobelli2011, Kushner2020}. Such prediction windows enable AID systems to preemptively modulate basal insulin delivery, issue hypoglycemia alarms, and assist patients in meal-time dosing decisions.

Despite substantial progress in applying deep learning to glucose forecasting~\citep{Fox2018, Li2020, Clara2022}, spanning recurrent, convolutional, and transformer-based architectures, two fundamental problems persist at long horizons. First, recursive autoregression compounds errors: a long short-term memory (LSTM) neural network that predicts $\hat{g}_{t+1}$ and feeds it back as input to predict $\hat{g}_{t+2}$ accumulates distributional shift over 48 5-minute steps (240 minutes), manifesting as a systematic negative bias (under-prediction) that increases monotonically across the forecasting horizon. Our experiments confirm that a recursive LSTM reaches a ME of $-$24.51 mg/dL at 240 minutes. Second, observational incompleteness: CGM signals alone do not reveal the current physiological state of a patient. Plasma insulin concentrations that have time-varying kinetics and dynamics \citep{Bergman1981}, and time-varying carbohydrate absorption in the gut \citep{DallaMan2007} are challenging for a neural network to model, yet they impact glucose dynamics for many hours in the future \citep{DallaMan2014}.

Mechanistic ordinary differential equation (ODE) models, such as the Hovorka system~\citep{Hovorka2004}, provide an effective way to model glucose dynamics using our understanding of how insulin and carbohydrates impact the dynamics. However, when parameterized with fixed or population-level parameters, these models are too rigid to capture time-varying (intra-patient) dynamics observed in real-world CGM data~\citep{Contreras2017}, and typically require patient-specific parameter estimation to address inter-patient variability, leading to degraded long-horizon forecasts. This work does not propose a fundamentally new deep learning backbone to compete with recent high-capacity models such as transformers but instead isolates and systematically reduces two specific failure modes within standard recurrent architectures.

To address the shortcomings of recursive autoregressive models and fixed-parameter ODE models, we propose PhysioSeq2Seq, a principled hybrid architecture whose core insight is that a patient-matched ODE digital twin need not produce the final prediction, but can instead supply internal physiological states as exogenous covariates to a neural sequence model.

Our proposed framework introduces three key novelties. First, rather than relying on a single fixed-parameter ODE or requiring expensive patient-specific calibration, we propose twin matching: a real-time, segment-level search over a pre-computed population (300 parameterizations~\citep{Hovorka2004} fitted to 307,474 CGM segments from 497 people with T1D in the T1DEXI study~\citep{Riddell2023}), enabling patient-specific physiological adaptation without retraining. Second, unlike prior hybrid models that use ODE outputs as predictions or residual targets, we inject all 10 internal ODE state variables, including remote insulin action and gut absorption states invisible to a data-only model, as exogenous covariates into both the encoder and decoder of a Seq2Seq LSTM, grounding the neural model in physiologically feasible dynamics across the full 7-hour window. Third, by adopting a simultaneous multi-step decoding strategy that produces all 48 future timesteps in a single forward pass, we structurally eliminate recursive error accumulation, the dominant source of long-horizon bias in autoregressive approaches, without requiring any architectural change to the underlying LSTM.

Evaluated on 74 held-out participants from the T1DEXI dataset, PhysioSeq2Seq achieves a MAE of 39.28~mg/dL and ME of $-$10.62~mg/dL at the 240-minute horizon, reducing 240-minute bias by 11.99~mg/dL over the recursive LSTM and outperforming all baselines across every forecast horizon.

While this work focuses on T1D glucose forecasting, the framework of matching a population of mechanistic ODE models and injecting their internal states into a Seq2Seq architecture explores a broader class of hybrid methods. We hypothesize that this pattern may extend to domains where a mechanistic ODE model exists but population-level parameters fail to capture individual variability. Future work will show how the method generalizes across multiple applications.


\section{Related Work}
\label{sec:related}
\subsection{Deep Learning for Glucose Forecasting}
Recurrent architectures, particularly LSTMs, have become the dominant data-driven approach to short-term glucose forecasting. \citet{Fox2018} showed that recursive autoregressive approaches suffer from compounding prediction error and that encoder--decoder architectures substantially reduce this accumulation across the full prediction horizon. \citet{Mirshekarian2019} found that LSTMs with meal and insulin features outperform SVR and expert baselines, though attention-based gains on synthetic data do not consistently transfer to free-living settings. \citet{Li2020} proposed convolutional recurrent neural networks that combine local feature extraction with temporal sequence modeling, achieving leading accuracy on both simulated and real patient data at 30- and 60-minute horizons. \citet{XieWang2020} provided a systematic benchmark of machine learning algorithms against classical autoregressive models for T1D glucose prediction, finding that model performance varies substantially across patients and horizons, with no single architecture consistently dominating, and that temporal convolutional networks show greater robustness to patient-level glucose variability than recurrent approaches. \citet{Clara2022} trained and evaluated LSTM models on a large T1D dataset of 250 patients and showed that prediction accuracy degrades with patient glucose variability. More recently, transformer-based architectures have been applied to multi-horizon CGM prediction. \citet{Karagoz2025} compared transformer variants for horizons up to 4 hours on T1D data, finding that patch-wise tokenization strategies outperform point-wise encoders by better capturing patterns in CGM signals. However, these purely data-driven models make no use of physiological prior knowledge and cannot provide mechanistic constraints on the predicted trajectory.

\subsection{Physics-Informed and Hybrid Modeling}
Mechanistic ODE models such as the glucose-insulin system of~\citet{Hovorka2004} describe glucose dynamics through interpretable compartmental equations, but population-level parameterizations cannot adapt to inter- and intra-person physiological variability in real-world data. \citet{Chen2018} introduced neural ODEs, showing that parameterizing the hidden-state derivative with a neural network yields a continuous-depth model trainable via adjoint-based backpropagation. \citet{Karniadakis2021} argued that embedding mechanistic structure into neural architectures through constrained loss functions or ODE-based inductive biases consistently improves generalization over purely data-driven approaches. \citet{Rackauckas2021} extended this with universal differential equations, which augment known ODE structures with neural residual terms to compensate for model mismatch. \citet{Zou2024} applied this paradigm to T1D on the T1DEXI dataset, pairing a mechanistic ODE with a neural component and a causal loss, demonstrating strong predictive performance on post-exercise glucose dynamics. In contrast, our method integrates ODE digital twin states as input features to a Seq2Seq LSTM, avoiding differentiable simulation while grounding the neural model in physiologically feasible dynamics.

\subsection{Digital Twins for Type 1 Diabetes}
The concept of a digital twin, a patient-specific virtual model calibrated from observed data, has attracted growing interest for T1D simulation and treatment optimization~\citep{Mosquera-Lopez2024}. \citet{CapponReplayBG2023} proposed ReplayBG, an open-source framework that identifies a personalized glucose-insulin model from CGM, meal, and insulin data to simulate counterfactual trajectories, demonstrating that CGM-based twinning alone suffices for personalization without invasive measurements. \citet{Cappon2025} reviewed eight T1D digital twin frameworks including ReplayBG, identifying the lack of real-world validation and cross-framework benchmarking as the most critical barriers to adoption. \citet{RoquemenEcheverri2026} proposed a neural network digital twin framework for T1D that embeds known physiological ODE constraints directly into the model architecture and learns patient-specific corrections on top of a shared population model. Their framework was validated on the same T1DEXI dataset used in this work, across 394 digital twins, producing glucose outcomes statistically equivalent to real observations in CGM time in range (70--180 mg/dL), time below range ($<70$ mg/dL), and time above range ($>180$ mg/dL). These results demonstrate that physiologically grounded neural network models can match real-world free-living T1D glucose dynamics.

This study is complementary: rather than using a neural twin as a simulator to reproduce full glucose trajectories, we use a classical ODE twin purely as a feature extractor, providing its 10 internal ODE state variables as exogenous covariates to a Seq2Seq LSTM encoder--decoder. This allows the neural model to learn residual dynamics that fixed-parameter population models cannot capture, without requiring the twin itself to be differentiable or end-to-end trainable.


\section{Proposed Approach}
\label{sec:proposed_approach}
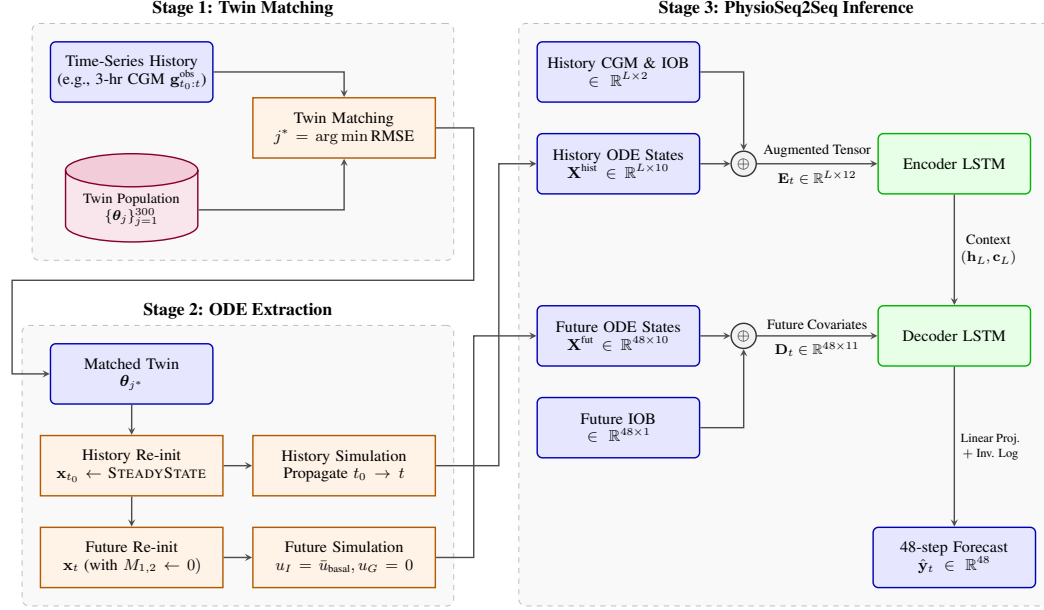
\begin{figure}[!htbp]
  \centering
  \resizebox{\textwidth}{!}{
    \begin{tikzpicture}[
        >=stealth,
        data/.style={rectangle, draw=blue!70!black, fill=blue!10, rounded corners=3pt, thick, align=center, text width=3.0cm, minimum height=1.2cm, font=\small},
        model/.style={rectangle, draw=green!70!black, fill=green!10, rounded corners=3pt, thick, align=center, text width=2.8cm, minimum height=1.2cm, font=\small},
        process/.style={rectangle, draw=orange!70!black, fill=orange!10, thick, align=center, text width=3.4cm, minimum height=1.2cm, font=\small},
        database/.style={cylinder, cylinder uses custom fill, cylinder body fill=purple!10, cylinder end fill=purple!20, shape border rotate=90, draw=purple!70!black, aspect=0.25, thick, align=center, text width=2.4cm, minimum height=1.5cm, font=\footnotesize},
        op/.style={circle, draw=black!80, fill=black!5, thick, inner sep=1pt, minimum size=0.5cm, font=\small},
        arrow/.style={->, thick, draw=black!70},
        stage/.style={draw=gray!50, dashed, rounded corners, inner sep=10pt, fill=gray!5},
        brace/.style={decorate, decoration={brace, amplitude=6pt}, thick, draw=gray!70},
        hovsys/.style={rectangle, draw=purple!70!black, fill=purple!10, rounded corners=4pt, thick, align=center, text width=3.2cm, minimum height=1.8cm, font=\scriptsize},
        dtbox/.style={rectangle, draw=purple!70!black, fill=purple!10, rounded corners=3pt, thick, align=center, font=\small},
        dtarrow/.style={->, thick, dashed, draw=purple!70!black},
        hovarrow/.style={->, thick, draw=purple!70!black},
        hovmodel/.style={draw=purple!40, dashed, rounded corners=8pt, thick, fill=purple!5, inner sep=15pt}
    ]

    \node[data] (hist_cgm) {Time-Series History\\{\small (e.g., 3-hr CGM $\mathbf{g}^{\text{obs}}_{t_0:t}$)}};
    \node[database, below=1.0cm of hist_cgm] (pop) {Twin Population\\$\{\boldsymbol{\theta}_j\}_{j=1}^{300}$};
    \path (hist_cgm) -- (pop) coordinate[midway] (mid1);
    \node[process, right=2.4cm of mid1] (matching) {Twin Matching\\$j^* = \arg\min \text{RMSE}$};

    \draw[arrow] (hist_cgm) -| (matching);
    \draw[arrow] (pop) -| (matching);

    \node[data, below=2cm of pop] (best_twin) {Matched Twin\\$\boldsymbol{\theta}_{j^*}$};

    \node[process, below=0.6cm of best_twin] (hist_reinit) {History Re-init\\$\mathbf{x}_{t_0} \leftarrow \textsc{SteadyState}$};
    \node[process, below=0.6cm of hist_reinit] (fut_reinit) {Future Re-init\\$\mathbf{x}_t \text{ (with } M_{1,2} \leftarrow 0)$};

    \node[process] (hist_prop) at (hist_reinit -| matching) {History Simulation\\Propagate $t_0 \to t$};
    \node[process] (fut_prop) at (fut_reinit -| matching) {Future Simulation\\$u_{I} = \bar{u}_{\text{basal}}, u_G = 0$};

    \path (pop.south) -- (best_twin.north) coordinate[midway] (gap_mid);
    \coordinate (wrap_left) at ([xshift=-0.75cm]best_twin.west);
    \draw[arrow] (matching.east) -- ++(0.75,0) |- ([yshift=0.25cm]gap_mid -| wrap_left) |- (best_twin.west);

    \draw[arrow] (best_twin) -- (hist_reinit);
    \draw[arrow] (hist_reinit) -- (hist_prop);
    \draw[arrow] (hist_reinit) -- (fut_reinit);
    \draw[arrow] (fut_reinit) -- (fut_prop);

    \path (matching.east |- fut_prop) coordinate (stage2_right);
    \coordinate (stage3_west) at ([xshift=2.0cm]stage2_right);

    \node[data, anchor=west] (hist_feat) at (stage3_west |- hist_cgm) {History CGM \& IOB\\$\in \mathbb{R}^{L \times 2}$};
    \node[data, below=0.6cm of hist_feat] (x_hist) {History ODE States\\$\mathbf{X}^{\text{hist}} \in \mathbb{R}^{L \times 10}$};

    \node[op, right=0.6cm of x_hist] (concat_hist) {$\oplus$};
    \node[model, right=2.4cm of concat_hist] (encoder) {Encoder LSTM};

    \draw[arrow] (x_hist) -- (concat_hist);
    \draw[arrow] (hist_feat) -| (concat_hist);
    \draw[arrow] (concat_hist) -- node[above, font=\footnotesize] {Augmented Tensor} node[below, font=\footnotesize] {$\mathbf{E}_t \in \mathbb{R}^{L \times 12}$} (encoder);

    \node[data, below=2.2cm of x_hist] (x_fut) {Future ODE States\\$\mathbf{X}^{\text{fut}} \in \mathbb{R}^{48 \times 10}$};
    \node[data, below=0.6cm of x_fut] (fut_feat) {Future IOB\\$\in \mathbb{R}^{48 \times 1}$};

    \node[op, right=0.6cm of x_fut] (concat_fut) {$\oplus$};
    \node[model] (decoder) at (concat_fut -| encoder) {Decoder LSTM};

    \draw[arrow] (x_fut) -- (concat_fut);
    \draw[arrow] (fut_feat) -| (concat_fut);
    \draw[arrow] (concat_fut) -- node[above, font=\footnotesize] {Future Covariates} node[below, font=\footnotesize] {$\mathbf{D}_t \in \mathbb{R}^{48 \times 11}$} (decoder);

    \draw[arrow] (encoder) -- node[right, font=\footnotesize, align=center] {Context\\$(\mathbf{h}_L, \mathbf{c}_L)$} (decoder);
    \draw[arrow] (hist_prop.east) -- ++(1.25, 0) |- (x_hist.west);
    \draw[arrow] (fut_prop.east) -- ++(0.75, 0) |- (x_fut.west);

    \node[data, below=3.185cm of decoder] (forecast) {48-step Forecast\\$\hat{\mathbf{y}}_t \in \mathbb{R}^{48}$};
    \draw[arrow] (decoder) -- node[right, font=\scriptsize, align=center] {Linear Proj.\\$+$ Inv. Log} (forecast);

    \begin{scope}[on background layer]
        \node[stage, fit=(hist_cgm) (pop) (matching)] (box1) {};
        \path (matching.east |- fut_prop.south) coordinate (box2_br);
        \node[stage, fit=(best_twin) (fut_reinit) (box2_br)] (box2) {};
        \node[stage, fit=(x_hist) (x_fut) (hist_feat) (fut_feat) (concat_hist) (concat_fut) (encoder) (decoder) (forecast)] (box3) {};
    \end{scope}

    \node[above=0cm of box1, font=\bfseries] {Stage 1: Twin Matching};
    \node[above=0cm of box2, font=\bfseries] {Stage 2: ODE Extraction};
    \node[above=0cm of box3, font=\bfseries] {Stage 3: PhysioSeq2Seq Inference};

\end{tikzpicture}
  }
  \caption{
    PhysioSeq2Seq end-to-end pipeline.
  }
  \label{fig:pipeline}
\end{figure}

PhysioSeq2Seq consists of three stages illustrated in Figure~\ref{fig:pipeline}. Stage~1 performs twin matching: for each input segment, the best-fitting ODE digital twin is selected from a pre-computed population of 300 parameterizations by minimizing RMSE over the 3-hour CGM history. Stage~2 performs ODE state extraction: the matched twin is propagated over both the history and the 4-hour forecast horizon to produce physiological state trajectories. Stage~3 performs PhysioSeq2Seq inference: those states are injected as exogenous covariates into a Seq2Seq encoder--decoder LSTM that produces all 48 forecast steps simultaneously. The following subsections detail each component.

\subsection{Problem Formulation}
\label{sec:problem}
We formalize the glucose forecasting task as a multi-step prediction problem over a fixed history window, where future covariates derived from the ODE digital twin are available to the decoder.

Let $\cgm_t \in [40, 400]$ mg/dL denote the CGM reading at discrete time $t$, sampled at $\Ts = 5$ minutes. Let $\xvec_t \in \mathbb{R}^{10}$ denote the full physiological state vector of the matched ODE digital twin (Section~\ref{sec:hovorka}) at time $t$. Let $\iob_t \in \mathbb{R}_{\geq 0}$ denote the insulin-on-board at time $t$, computed by aggregating all insulin delivered more than 60 minutes prior to $t$ via a pharmacokinetic activity model (Section~\ref{sec:iob}).

Given a history window of up to $L = 37$ steps (36 steps of history $+$ 1 step for the decision time) ending at decision time $t$, with shorter sequences zero-padded to the full length:
\[
  \mathcal{H}_t = \bigl\{(\cgm_k,\; \iob_k,\; \xvec_k)\bigr\}_{k=t-L+1}^{t},
\]
predict the future CGM trajectory over $H = 48$ steps (4 hours):
\[
  \hat{\yvec}_t = (\hat{\cgm}_{t+1},\; \hat{\cgm}_{t+2},\;
  \ldots,\; \hat{\cgm}_{t+H}) \in \mathbb{R}^H.
\]
Unlike autoregressive formulations, $\hat{\mathbf{y}}_t$ is produced in a single forward pass, conditioned on future exogenous signals derived from the digital twin (Section~\ref{sec:twin_matching}).
\subsection{The ODE Digital Twin Model}
\label{sec:hovorka}

The ODE digital twin models glucose-insulin dynamics in T1D through a 10-state compartmental system~\citep{Hovorka2004}, providing physiologically interpretable state variables that serve as covariates for the Seq2Seq model. The state vector $\xvec_k$ is defined as:

\begin{equation}
  \xvec_k = \bigl[
  \; \underbrace{Q_1,\; Q_2}_{\text{glucose}},\;\;
  \underbrace{S_1,\; S_2,\; I}_{\text{insulin PK}},\;\;
  \underbrace{X_1,\; X_2,\; X_3}_{\text{remote action}},\;\;
  \underbrace{M_1,\; M_2}_{\text{gut absorption}}
  \bigr]_k^\top \in \mathbb{R}^{10}.
\end{equation}
The glucose compartments $Q_1$ and $Q_2$ represent accessible and non-accessible glucose mass (mmol/kg). States $S_1$, $S_2$, and $I$ model subcutaneous insulin absorption delay and the resulting plasma concentration (mU/L). The remote action states $X_1$, $X_2$, $X_3$ modulate glucose disposal, distribution, and endogenous production, while $M_1$ and $M_2$ govern gut carbohydrate absorption (mmol/kg). At the $\Ts = 5$-minute sampling interval, the model advances one step forward via a discrete-time state transition:
\begin{equation}
  \xvec_{k+1} = A_d(\xvec_k)\xvec_k + B_d\,u_{I,k} + D_d(\xvec_k) + G_d\,u_{G,k}.
  \label{eq:hovorka_discrete}
\end{equation}
Here $A_d(\xvec_k)\xvec_k$ evolves the physiological state, $B_d\,u_{I,k}$ adds insulin delivery (mU/kg/min), $D_d(\xvec_k)$ captures nonlinear processes such as glucose disposal and hepatic production, and $G_d\,u_{G,k}$ adds meal carbohydrate absorption (g/min). The four matrices are obtained by exponentiating the augmented continuous-time system matrix over $T_s$:
\begin{equation}
  e^{\,\bigl[\,A_p \;\big|\; B_p \;\big|\; D_p \;\big|\; G_p\,\bigr]\,\Ts} \;\;\longrightarrow\;\; \bigl[\,A_d \;\big|\; B_d \;\big|\; D_d \;\big|\; G_d\,\bigr],
  \label{eq:zoh}
\end{equation}
where $A_p(\mathbf{x}_k) \in \mathbb{R}^{10 \times 10}$ is the state-dependent system matrix, $B_p = \mathbf{e}_3$ is the insulin input vector, $D_p(\mathbf{x}_k)$ captures nonlinear state-dependent terms such as glucose disposal and hepatic glucose production, and $G_p = \mathbf{e}_9$ is the meal input vector (Appendix~\ref{app:ode_full}). Finally, the CGM reading (mg/dL) is recovered from the accessible glucose compartment $Q_1$ by a simple unit conversion:
\begin{equation}
  \cgm_k = \frac{18\,Q_{1,k}}{V_G},
  \label{eq:cgm_output}
\end{equation}
where $V_G$ (L/kg) is the patient-specific glucose distribution volume.

\subsection{Insulin-on-Board used in Seq2Seq Model}
\label{sec:iob}
To capture the lagged metabolic effect of past insulin delivery, we compute insulin-on-board (IOB) as an exogenous feature for both the encoder and decoder inputs of the Seq2Seq LSTM, and for the recursive LSTM baseline. IOB quantifies the total insulin still metabolically active at time $t$. Although the ODE state $I$ (Section~\ref{sec:hovorka}) encodes plasma insulin concentration, we compute IOB separately via a closed-form pharmacokinetic approximation following prior work~\citep{Clara2022}. This approximation is computed as:
\begin{equation}
  \iob_t = \sum_{e \in \mathcal{E}_{<t}} \text{dose}(e)\cdot \varphi\left(t_{\text{ref}} - t_e\right),
\end{equation}
where $\mathcal{E}_{<t}$ is the set of all insulin delivery events prior to $t$, $t_e$ is the delivery timestamp of event $e$ (in minutes), and $t_{\text{ref}} = t - 60\;\text{min}$ is an effective reference time that excludes insulin delivered within the past 60 minutes, since it has not yet reached systemic circulation. The activity function $\varphi(\Delta)$ models the three-phase subcutaneous absorption curve, whereby the variable $\Delta$ is the time difference in minutes between t and an insulin dosed in the past:
\begin{equation}
  \varphi(\Delta) =
  \begin{cases}
    \Delta / 30             & \Delta \leq 30\;\text{min}      \\
    1.0                     & 30 < \Delta \leq 90\;\text{min} \\
    e^{-0.012(\Delta - 90)} & \Delta > 90\;\text{min},
  \end{cases}
\end{equation}
Basal insulin at rate $r_{\text{U/hr}}$ is decomposed into consecutive 5-minute micro-doses of $r_{\text{U/hr}}/12$\,U each. Pump suspensions (rate~$= 0$) are ignored. Extended boluses (duration ${>}\,5$ min) are spread uniformly into $\lfloor d_{\min}/5 \rfloor$ steps of $\text{dose}/\lfloor d_{\min}/5 \rfloor$\,U each, and manual boluses are treated as single instantaneous events. The raw IOB value is then linearly normalized to $[-1, 1]$ using a fixed physiological reference range $[0,4.53]$ U:
\begin{equation}
  \widetilde{\iob}_t = 2 \cdot \frac{\min(\iob_t,\; 4.53)}{4.53} - 1, \qquad \iob_t \geq 0.
\end{equation}

\subsection{Twin Matching: Patient-Specific State Identification}
\label{sec:twin_matching}
To adapt the ODE digital twin to each patient segment, we select the best-matching parameterization from a pre-computed population by minimizing simulated CGM error over the observed history.

A pre-computed population of $J = 300$ virtual patients provides distinct parameterizations $\{\boldsymbol{\theta}_j\}_{j=1}^{300}$ of the ODE digital twin, each with a different insulin sensitivity. For each input segment, we identify the best-fitting twin by minimizing the RMSE between the simulated and observed CGM over the 3-hour history window:
\begin{equation}
  j^* = \underset{j \in \{1,\ldots,300\}}{\arg\min} \sqrt{\frac{1}{L}\sum_{k=1}^{L} \bigl(\hat{g}_k^{(j)} - g_k^{\text{obs}}\bigr)^2}.
\end{equation}
Here $\hat{\cgm}_k^{(j)}$ is the simulated glucose of twin $j$ initialized at the observed CGM and glucose slope at $t_0$.

The twin initialization uses a closed-form steady-state ODE solver that sets the $Q_1$ and $\dot{Q}_1$ states to the glucose and glucose slope at the start of the window, while setting all other state variable derivatives to zero. We then rerun the selected twin $j^*$ over the same 3-hour history with state logging enabled to recover the hidden state trajectory $\mathbf{X}^{\mathrm{hist}} \in \mathbb{R}^{L \times 10}$. This second pass is required because the matching pass only returns the best-fitting twin, not its intermediate ODE states.

At the decision time $t$, we reinitialize the selected twin for forecasting using the observed CGM value $g_t$ and estimated CGM slope $\dot{g}_t$. The steady-state solver anchors $Q_1$ to the observed CGM, restores $(S_1, S_2)$ from the history pass terminal state, and sets meal compartments to zero under the basal-only assumption. Finally, we propagate the reinitialized twin forward for $H=48$ five-minute steps using constant basal delivery and zero future bolus/carbohydrate inputs, producing the future state trajectory $\mathbf{X}^{\mathrm{fut}} \in \mathbb{R}^{H \times 10}$ (Appendix~\ref{app:algorithm}).

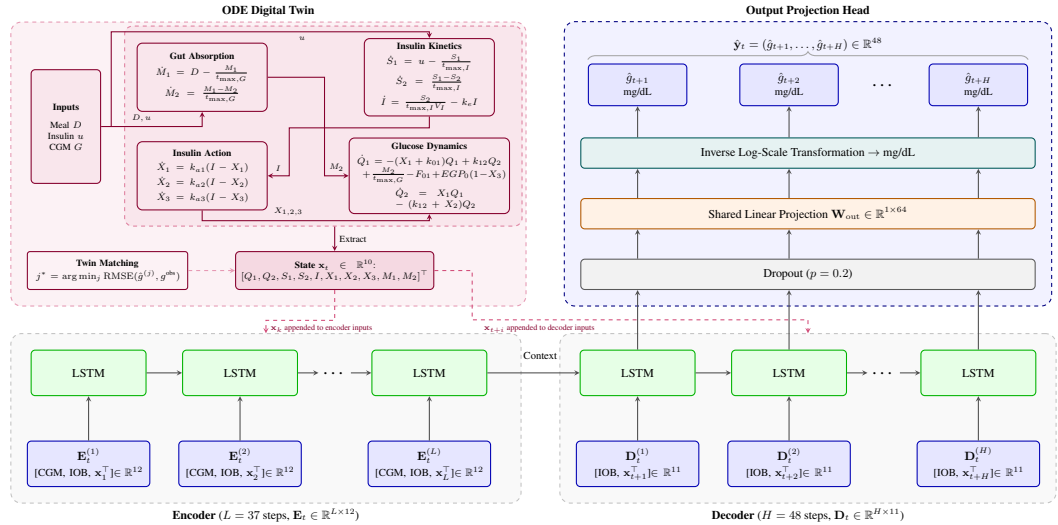
\begin{figure}[!htbp]
  \centering
  \resizebox{\textwidth}{!}{
    \begin{tikzpicture}[
        >=stealth,
        data/.style={rectangle, draw=blue!70!black, fill=blue!10, rounded corners=3pt, thick, align=center, text width=3.0cm, minimum height=1.2cm, font=\small},
        model/.style={rectangle, draw=green!70!black, fill=green!10, rounded corners=3pt, thick, align=center, text width=2.8cm, minimum height=1.2cm, font=\small},
        process/.style={rectangle, draw=orange!70!black, fill=orange!10, thick, align=center, text width=3.4cm, minimum height=1.2cm, font=\small},
        database/.style={cylinder, cylinder uses custom fill, cylinder body fill=purple!10, cylinder end fill=purple!20, shape border rotate=90, draw=purple!70!black, aspect=0.25, thick, align=center, text width=2.4cm, minimum height=1.5cm, font=\footnotesize},
        op/.style={circle, draw=black!80, fill=black!5, thick, inner sep=1pt, minimum size=0.5cm, font=\small},
        arrow/.style={->, thick, draw=black!70},
        brace/.style={decorate, decoration={brace, amplitude=6pt}, thick, draw=gray!70},
        stage/.style={draw=gray!50, dashed, rounded corners=8pt, inner sep=12pt, fill=gray!5},
        hovmodel/.style={draw=purple!40, dashed, rounded corners=8pt, thick, fill=purple!5, inner sep=12pt},
        hovsys/.style={rectangle, draw=purple!70!black, fill=purple!10, rounded corners=4pt, thick, align=center, text width=3.2cm, minimum height=1.8cm, font=\scriptsize},
        dtbox/.style={rectangle, draw=purple!70!black, fill=purple!10, rounded corners=3pt, thick, align=center, font=\small},
        dtarrow/.style={->, thick, dashed, draw=purple!70!black},
        hovarrow/.style={->, thick, draw=purple!70!black}
    ]

    \node[model] (e1) at (0, 0) {LSTM};
    \node[model] (e2) at (4.0, 0) {LSTM};
    \node[font=\Large] (edots) at (6.5, 0) {\dots};
    \node[model] (eL) at (9.0, 0) {LSTM};
    \draw[arrow] (e1)--(e2); \draw[arrow] (e2)--(edots); \draw[arrow] (edots)--(eL);

    \node[data] (i1) at (0, -2.4) {$\mathbf{E}_t^{(1)}$\\[2pt] {\footnotesize[CGM,\ IOB,\ $\mathbf{x}_1^\top$]}{\footnotesize$\in\mathbb{R}^{12}$}};
    \node[data] (i2) at (4.0, -2.4) {$\mathbf{E}_t^{(2)}$\\[2pt]{\footnotesize[CGM,\ IOB,\ $\mathbf{x}_2^\top$]}{\footnotesize$\in\mathbb{R}^{12}$}};
    \node[data] (iL) at (9.0, -2.4) {$\mathbf{E}_t^{(L)}$\\[2pt]{\footnotesize[CGM,\ IOB,\ $\mathbf{x}_L^\top$]}{\footnotesize$\in\mathbb{R}^{12}$}};
    \draw[arrow] (i1)--(e1); \draw[arrow] (i2)--(e2); \draw[arrow] (iL)--(eL);

    \node[model] (d1) at (14.5, 0) {LSTM};
    \node[model] (d2) at (18.5, 0) {LSTM};
    \node[font=\Large] (ddots) at (21.0, 0) {\dots};
    \node[model] (dH) at (23.5, 0) {LSTM};
    \draw[arrow] (d1)--(d2); \draw[arrow] (d2)--(ddots); \draw[arrow] (ddots)--(dH);
    \draw[arrow] (eL) -- node[pos=0.575, above=0.22cm, font=\footnotesize, align=center]{Context} (d1);

    \node[data] (f1) at (14.5, -2.4) {$\mathbf{D}_t^{(1)}$\\[2pt]{\footnotesize[IOB,\ $\mathbf{x}_{t+1}^\top$]}{\footnotesize$\in\mathbb{R}^{11}$}};
    \node[data] (f2) at (18.5, -2.4) {$\mathbf{D}_t^{(2)}$\\[2pt] {\footnotesize[IOB,\ $\mathbf{x}_{t+2}^\top$]}{\footnotesize$\in\mathbb{R}^{11}$}};
    \node[data] (fH) at (23.5, -2.4) {$\mathbf{D}_t^{(H)}$\\[2pt] {\footnotesize[IOB,\ $\mathbf{x}_{t+H}^\top$]}{\footnotesize$\in\mathbb{R}^{11}$}};
    \draw[arrow] (f1)--(d1); \draw[arrow] (f2)--(d2); \draw[arrow] (fH)--(dH);

    \node[rectangle, draw=gray!70!black, fill=gray!10, rounded corners=3pt, thick, align=center, minimum width=12.0cm, minimum height=0.8cm, font=\small] (dropout) at (19.0, 2.6) {Dropout ($p=0.2$)};

    \node[rectangle, draw=orange!70!black, fill=orange!10, rounded corners=3pt, thick, align=center, minimum width=12.0cm, minimum height=0.8cm, font=\small] (dense) at (19.0, 4.2) {Shared Linear Projection $\mathbf{W}_{\mathrm{out}}\in\mathbb{R}^{1\times64}$};

    \node[rectangle, draw=teal!70!black, fill=teal!10, rounded corners=3pt, thick, align=center, minimum width=12.0cm, minimum height=0.8cm, font=\small] (invlog) at (19.0, 5.8) {Inverse Log-Scale Transformation $\rightarrow$ mg/dL};

    \node[rectangle, draw=blue!70!black, fill=blue!10, rounded corners=3pt, thick, align=center, minimum width=2.6cm, minimum height=1.1cm, font=\small] (o1) at (14.5, 7.6) {$\hat{g}_{t+1}$\\{\footnotesize mg/dL}};
    \node[rectangle, draw=blue!70!black, fill=blue!10, rounded corners=3pt, thick, align=center, minimum width=2.6cm, minimum height=1.1cm, font=\small] (o2) at (18.5, 7.6) {$\hat{g}_{t+2}$\\{\footnotesize mg/dL}};
    \node[font=\Large] at (21.0, 7.6) {\dots};
    \node[rectangle, draw=blue!70!black, fill=blue!10, rounded corners=3pt, thick, align=center, minimum width=2.6cm, minimum height=1.1cm, font=\small] (oH) at (23.5, 7.6) {$\hat{g}_{t+H}$\\{\footnotesize mg/dL}};

    \draw[arrow] (d1)--(d1 |- dropout.south);
    \draw[arrow] (d2)--(d2 |- dropout.south);
    \draw[arrow] (dH)--(dH |- dropout.south);

    \draw[arrow] (d1 |- dropout.north)--(d1 |- dense.south);
    \draw[arrow] (d2 |- dropout.north)--(d2 |- dense.south);
    \draw[arrow] (dH |- dropout.north)--(dH |- dense.south);

    \draw[arrow] (d1 |- dense.north)--(d1 |- invlog.south);
    \draw[arrow] (d2 |- dense.north)--(d2 |- invlog.south);
    \draw[arrow] (dH |- dense.north)--(dH |- invlog.south);

    \draw[arrow] (d1 |- invlog.north)--(o1);
    \draw[arrow] (d2 |- invlog.north)--(o2);
    \draw[arrow] (dH |- invlog.north)--(oH);

    \draw[brace] ($(o1.north west)+(0,0.1)$) -- ($(oH.north east)+(0,0.1)$);
    \node[font=\small\bfseries, align=center, inner sep=0pt] (brace_text) at ($(o1.north west)!0.5!(oH.north east)+(0,0.5)$) {$\hat{\mathbf{y}}_t = (\hat{g}_{t+1}, \ldots, \hat{g}_{t+H}) \in \mathbb{R}^{48}$};

    \node[rectangle, draw=purple!70!black, fill=purple!10, rounded corners=3pt, thick, align=center, text width=1.6cm, minimum height=3.0cm, font=\scriptsize]
    (inp) at (-0.6, 6.5) {\textbf{Inputs}\\[6pt]Meal $D$\\Insulin $u$\\CGM $G$};

    \node[hovsys] (gut)     at (3.0, 7.8)
    {\textbf{Gut Absorption}\\[3pt]
        $\dot{M}_1 = D - \tfrac{M_1}{t_{\max,G}}$\\[2pt]
        $\dot{M}_2 = \tfrac{M_1-M_2}{t_{\max,G}}$};

    \node[hovsys] (ins_kin) at (9.0, 7.8)
    {\textbf{Insulin Kinetics}\\[3pt]
        $\dot{S}_1 = u - \tfrac{S_1}{t_{\max,I}}$\\[2pt]
        $\dot{S}_2 = \tfrac{S_1-S_2}{t_{\max,I}}$\\[2pt]
        $\dot{I} = \tfrac{S_2}{t_{\max,I}V_I} - k_e I$};

    \node[hovsys] (ins_act) at (3.0, 5.2)
    {\textbf{Insulin Action}\\[3pt]
        $\dot{X}_1 = k_{a1}(I - X_1)$\\[2pt]
        $\dot{X}_2 = k_{a2}(I - X_2)$\\[2pt]
        $\dot{X}_3 = k_{a3}(I - X_3)$};

    \node[hovsys, text width=4.0cm] (gluc) at (9.0, 5.2)
    {\textbf{Glucose Dynamics}\\[3pt]
        $\dot{Q}_1 = -(X_1 + k_{01})Q_1 + k_{12}Q_2$\\
        $\phantom{\dot{Q}}+\tfrac{M_2}{t_{\max,G}} - F_{01} + EGP_0(1-X_3)$\\[2pt]
        $\dot{Q}_2 = X_1 Q_1$\\
        $\phantom{\dot{Q}}-(k_{12}+X_2)Q_2$};

    \node[rectangle, draw=purple!70!black, fill=purple!15, rounded corners=3pt, thick, align=center, text width=5.0cm, minimum height=1.0cm, font=\scriptsize]
    (state_out) at (6.5, 2.7)
    {\textbf{State} $\mathbf{x}_t\!\in\!\mathbb{R}^{10}$:\quad$[Q_1,Q_2,S_1,S_2,I,X_1,X_2,X_3,M_1,M_2]^\top$};

    \node[rectangle, draw=purple!70!black, fill=purple!5, rounded corners=2pt, font=\scriptsize, thick, align=center, text width=4.0cm, minimum height=1.0cm] (match_note) at (0.5, 2.7) {\textbf{Twin Matching}\\$j^*=\arg\min_j \mathrm{RMSE}(\hat{g}^{(j)}, g^{\text{obs}})$};

    \draw[hovarrow] (inp.east) -| node[above,pos=0.2,font=\tiny]{$D,u$} coordinate[pos=0.05] (branchpt) (gut.south);
    \draw[hovarrow] (branchpt) -- ++(0,2.5) -| (ins_kin.north) node[pos=0.3,below,font=\tiny]{$u$};
    \draw[hovarrow] (ins_kin.south) -- ++(0,-0.3) -- ++(-3.7,0) |- node[above,pos=0.75,font=\tiny]{$I$} (ins_act.east);
    \draw[hovarrow] (gut.east) -- ++(1.5,0) -- ++(0,-2.6) -- node[above,font=\tiny]{$M_2$} (gluc.west);
    \draw[hovarrow] (ins_act.south) -- ++(0,-0.27) -- ++(2.9,0) -| node[above,pos=-0.1,font=\tiny]{$X_{1,2,3}$} (gluc.south);

    \draw[->,draw=purple!40,dashed,thick] (match_note.east) -- (state_out.west);

    \begin{scope}[on background layer]
        \coordinate (L_W) at (-1.5, 0);
        \coordinate (L_E) at (11.0, 0);
        \coordinate (R_W) at (13.0, 0);
        \coordinate (R_E) at (25.0, 0);
        \node[hovmodel, fit=(inp)(gut)(ins_kin)(ins_act)(gluc)(match_note)(state_out) (L_W |- state_out.center) (L_E |- state_out.center)] (hov_box) {};
        \node[draw=purple!50, dashed, rounded corners=6pt, thick, fill=purple!10, inner sep=9pt, fit=(gut)(ins_kin)(ins_act)(gluc)] (physio_box) {};
        \node[stage, fit=(e1)(eL)(i1)(iL) (L_W |- e1.center) (L_E |- e1.center)] (enc_box) {};
        \node[draw=blue!50!black, dashed, rounded corners=8pt, thick, fill=blue!5, inner sep=12pt, fit=(dropout)(dense)(invlog)(o1)(o2)(oH)(brace_text) (R_W |- dense.center) (R_E |- dense.center)] (out_box) {};
        \node[stage, fit=(d1)(dH)(f1)(fH) (R_W |- d1.center) (R_E |- d1.center)] (dec_box) {};
    \end{scope}

    \node[above=0cm of hov_box, font=\bfseries\small]{ODE Digital Twin};
    \node[above=0cm of out_box, font=\bfseries\small]{Output Projection Head};
    \node[below=0cm of enc_box, font=\small]{\textbf{Encoder} ($L=37$ steps, $\mathbf{E}_t \in \mathbb{R}^{L \times 12}$)};
    \node[below=0cm of dec_box, font=\small]{\textbf{Decoder} ($H=48$ steps, $\mathbf{D}_t \in \mathbb{R}^{H \times 11}$)};

    \draw[->, thick, dashed, draw=purple!70] (state_out.south) -- ++(0,-0.8) -| node[right, font=\tiny, color=purple!70!black, pos=0.8, align=center] {$\mathbf{x}_{k}$ appended to encoder inputs} (enc_box.north);
    \draw[->, thick, dashed, draw=purple!70] (state_out.east) -- ++(1,0) -- ++(0,-1.3) -| node[below, font=\tiny, color=purple!70!black, pos=0.1]{$\mathbf{x}_{t+i}$ appended to decoder inputs} (dec_box.north);
    \draw[hovarrow] (physio_box.south -| state_out.north) -- node[right, font=\scriptsize, align=left] {Extract} (state_out.north);

\end{tikzpicture}
  }
  \caption{PhysioSeq2Seq architecture.}
  \label{fig:architecture}
\end{figure}
\subsection{PhysioSeq2Seq: Hybrid Encoder--Decoder Architecture}
\label{sec:seq2seq}
Figure~\ref{fig:architecture} illustrates the PhysioSeq2Seq architecture used for CGM prediction. Generally, the encoder compresses an augmented history (observed time series, optional exogenous inputs, ODE internal states) into a fixed context vector. The decoder uses this context plus future ODE state projections to produce the full forecast horizon simultaneously, eliminating recursive error accumulation.

\paragraph{Feature construction.}
The encoder input tensor augments the CGM and IOB history with the matched ODE states:
\begin{equation}
  \mathbf{E}_t^{(k)} = \Bigl[\underbrace{\tilde{g}_k}_{\text{scaled CGM}}\quad,\quad \underbrace{\widetilde{\iob}_k}_{\text{scaled IOB}}\quad,\quad  \underbrace{\xvec_k^\top}_{10\;\text{ODE states}}\Bigr] \in \mathbb{R}^{12}, \quad k = 1, \ldots, L,
\end{equation}
forming $\mathbf{E}_t \in \mathbb{R}^{L \times 12}$. The decoder input tensor includes future exogenous estimates for IOB and ODE states, with CGM after the prediction time excluded to prevent leakage of future ground truth:
\begin{equation}
  \label{eq:decoder_input}
  \mathbf{D}_t^{(i)} = \Bigl[\quad\underbrace{\widetilde{\iob}_{t+i}}_{\text{future IOB}}\quad,\quad \underbrace{\xvec_{t+i}^\top}_{10\;\text{ODE states}}\Bigr] \in \mathbb{R}^{11}, \quad i = 1, \ldots, H,
\end{equation}
forming $\mathbf{D}_t \in \mathbb{R}^{H \times 11}$. CGM is also log-scaled to $[-1, 1]$:
\begin{equation}
  \label{eq:log}
  \tilde{g} = 2 \cdot \frac{\log g - \log 40}{\log 400 - \log 40} - 1,
\end{equation}
and IOB is normalized as described in Section~\ref{sec:iob}.

\paragraph{Encoder.}
A single-layer LSTM with $d = 64$ hidden units processes the full $L$-step augmented history sequence $\mathbf{E}_t \in \mathbb{R}^{L \times 12}$:

\begin{equation}
  \bigl(\hvec_L^{\text{enc}},\; \cvec_L^{\text{enc}}\bigr) = \mathrm{LSTM}_{\mathrm{enc}}\bigl( \mathbf{E}_t;\; \mathbf{W}_{\mathrm{enc}} \bigr),
\end{equation}

where $\mathbf{W}_{\mathrm{enc}} \in \mathbb{R}^{4d \times (12 + d)}$ are the encoder LSTM weights. The output is the pair $(\hvec_L^{\text{enc}}, \cvec_L^{\text{enc}}) \in \mathbb{R}^{64} \times \mathbb{R}^{64}$, encoding the complete physiological history into a fixed-dimensional context. Only the final hidden and cell states are retained, and all intermediate hidden states are discarded.

\paragraph{Decoder.}
The decoder LSTM is initialized with the encoder's final states $(\mathbf{h}_L^{\text{enc}}, \mathbf{c}_L^{\text{enc}})$ and processes $\mathbf{D}_t \in \mathbb{R}^{H \times 11}$. The decoder returns the full sequence of $H$ hidden states $\mathbf{O} = (\mathbf{O}_1, \ldots, \mathbf{O}_H)$. A dropout layer with $p = 0.2$ is applied to $\mathbf{O}$ during training to regularize the output sequence.
\paragraph{Output projection.}
A shared linear layer maps each decoder timestep independently to a single scaled CGM prediction:
\begin{equation}
  \hat{\tilde{g}}_{t+i} = \mathbf{W}_{\mathrm{out}}\,\mathbf{O}_i+ b_{\mathrm{out}}, \quad i = 1, \ldots, H,
  \label{eq:output_proj}
\end{equation}
where $\mathbf{W}_{\mathrm{out}} \in \mathbb{R}^{1 \times 64}$ and the activation is linear. The scaled predictions are mapped back to mg/dL by inverting Equation~(\ref{eq:log}):
\begin{equation}
  \hat{g}_{t+i} = \exp\Bigl(\tfrac{\hat{\tilde{g}}_{t+i} + 1}{2} \cdot (\log 400 - \log 40) + \log 40\Bigr).
  \label{eq:inv_transform}
\end{equation}
The complete 48-step trajectory $\hat{\yvec}_t = (\hat{g}_{t+1}, \ldots, \hat{g}_{t+H})$ is produced in a single forward pass. No predicted CGM value is ever fed back as input to the model, eliminating the recursive error accumulation that afflicts autoregressive approaches.
\paragraph{Loss function.}
Training minimizes the Huber loss computed in scaled space over all 48 future steps, with $\delta = 1.0$:
\begin{equation}
  \mathcal{L}(\hat{\yvec}, \yvec) = \frac{1}{H} \sum_{i=1}^{H} \ell_\delta\bigl( \hat{\tilde{g}}_{t+i} - \tilde{g}_{t+i} \bigr), \quad \ell_\delta(r) =
  \begin{cases}
    \tfrac{1}{2}r^2,                             & |r| \leq \delta \\
    \delta\left(|r| - \tfrac{1}{2}\delta\right), & |r| > \delta,
  \end{cases}
\end{equation}

\paragraph{Training protocol.}
The model is trained using Adam~\citep{Kingma2015} with $\eta_0 = 10^{-3}$, $\beta_1 = 0.9$, $\beta_2 = 0.999$, a batch size of 64, and up to 100 epochs. Early stopping with patience 15 is applied based on validation Huber loss, and the learning rate is reduced by a factor of $0.5$ after 5 epochs without improvement, down to a minimum of $\eta_{\min} = 10^{-5}$.


\section{Experiments}
\label{sec:experiments}
\subsection{Dataset}
\label{sec:dataset}
We use the T1DEXI dataset~\citep{Riddell2023}, a large-scale dataset of individuals with type~1 diabetes equipped with CGM devices and insulin pumps. Each segment is a 7-hour window (3-hour history + 4-hour forecast) sampled at 5-minute intervals, yielding a 37-step encoder input and a 48-step decoder target. Segments are filtered according to the dataset quality control pipeline described in Appendix~\ref{app:qc}. Data are partitioned patient by patient using a fixed random seed into training (70\%), validation (15\%), and test (15\%) sets, ensuring no individual appears in more than one split. The shared test set of 41,480 segments is the intersection of segments evaluable by all models. Table~\ref{tab:dataset} summarizes the dataset statistics.

\begin{table}[t]
  \caption{T1DEXI dataset statistics for the glucose forecasting task.}
  \label{tab:dataset}
  \centering
  \small
  \begin{tabular}{lrrrrrr}
    \toprule
    \multicolumn{1}{c}{\textbf{Metric}}     &
    \multicolumn{1}{c}{\textbf{Total}}      &
    \multicolumn{1}{c}{\textbf{Extracted}}  &
    \multicolumn{1}{c}{\textbf{Training}}   &
    \multicolumn{1}{c}{\textbf{Validation}} &
    \multicolumn{1}{c}{\textbf{Test}}       &
    \multicolumn{1}{c}{\textbf{Shared Test}}                                               \\
    \midrule
    Number of Patients                      &
    497                                     & 497     & 348     & 75     & 74     & 74     \\
    Number of Segments                      &
    307,474                                 & 289,859 & 203,935 & 44,112 & 44,501 & 41,480 \\
    \bottomrule
  \end{tabular}
\end{table}
\begin{figure}[t]
  \centering
  \includegraphics[width=\linewidth]{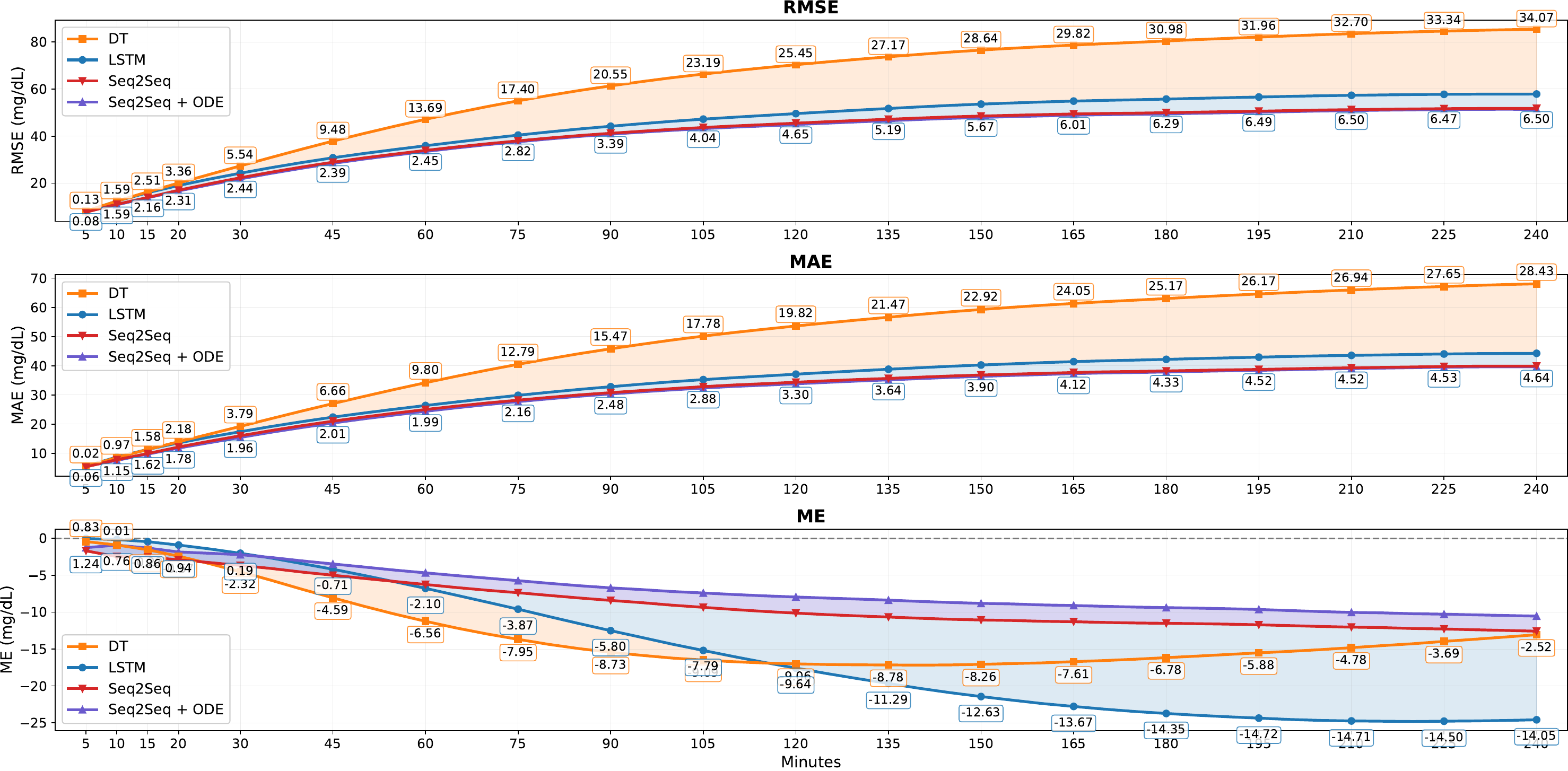}
  \caption{Model performance comparison across horizons from 5 to 240 minutes.}
  \label{fig:mae_me}
\end{figure}

\subsection{Baselines}
\label{sec:baselines}
We compare PhysioSeq2Seq against three baselines. The Digital Twin (DT) uses the ODE model with best-matched twin parameters under basal-only conditions to forecast glucose. The recursive LSTM is a single-layer LSTM (300 units) trained on CGM and IOB using an expanding-window strategy where the input grows by one step per prediction extending out to the full 4-hour prediction horizon. The Seq2Seq $-$ ODE uses the identical encoder--decoder architecture as PhysioSeq2Seq but with only CGM and IOB features and excluding the ODE state variable features. We also evaluate a naive baseline that predicts all future glucose values as the last observed measurement ($g_{t+h}=g_t$).

\subsection{Results}
\label{sec:results}
We evaluate performance using root mean squared error (RMSE), mean absolute error (MAE), and mean error (ME). Table~\ref{tab:main_results} summarizes results at selected forecast horizons.

\begin{table}[!htbp]
  \setlength{\tabcolsep}{2pt}
  \caption{Model performance at selected forecast horizons on 41,480 shared test segments (74 patients).}
  \label{tab:main_results}
  \centering
  \small
  \begin{tabular}{lcccccccccccc}
    \toprule
                                         & \multicolumn{3}{c}{\textbf{30 min}} &
    \multicolumn{3}{c}{\textbf{60 min}}  &
    \multicolumn{3}{c}{\textbf{120 min}} &
    \multicolumn{3}{c}{\textbf{240 min}}                                                                          \\
    \cmidrule(lr){2-4}\cmidrule(lr){5-7}
    \cmidrule(lr){8-10}\cmidrule(lr){11-13}
    \multicolumn{1}{c}{\textbf{Model}}   &
    RMSE                                 & MAE                                 & ME             &
    RMSE                                 & MAE                                 & ME             &
    RMSE                                 & MAE                                 & ME             &
    RMSE                                 & MAE                                 & ME                               \\
    \midrule
    Naive Baseline                       & 26.11                               & 18.30          & \textbf{0.10} &
    39.76                                & 28.55                               & \textbf{0.21}  &
    56.19                                & 41.26                               & \textbf{0.35}  &
    69.37                                & 51.55                               & \textbf{0.34}                    \\
    Digital Twin                         & 27.65                               & 19.11          & $-$4.67       &
    48.02                                & 34.19                               & $-$11.66       &
    71.96                                & 53.72                               & $-$17.84       &
    87.60                                & 67.90                               & $-$14.30                         \\
    Recursive LSTM                       & 24.77                               & 17.27          & $-$2.03       &
    36.82                                & 26.26                               & $-$6.79        &
    51.57                                & 37.05                               & $-$17.59       &
    61.51                                & 44.11                               & $-$24.51                         \\
    Seq2Seq $-$ ODE                      & 22.46                               & 15.71          & $-$3.62       &
    34.46                                & 24.64                               & $-$6.22        &
    47.20                                & 34.00                               & $-$10.08       &
    54.84                                & 39.54                               & $-$12.52                         \\
    Seq2Seq $+$ ODE                      & \textbf{21.71}                      & \textbf{15.12} & $-$2.21       &
    \textbf{33.71}                       & \textbf{24.00}                      & $-$4.73        &
    \textbf{46.16}                       & \textbf{33.40}                      & $-$8.04        &
    \textbf{54.12}                       & \textbf{39.28}                      & $-$10.62                         \\
    \bottomrule
  \end{tabular}
\end{table}

The results reveal three key patterns. First, the ODE model performs the worst compared with other models, indicating that ODE-based models, even when personalized with a best-matching digital twin, perform worse at modeling glucose dynamics than data-driven models. Second, the Seq2Seq architecture drives the majority of improvement over the ODE model and the recursive model. Replacing recursive autoregression with a single-pass decoder reduces 240-minute ME from $-$24.51 to $-$12.52~mg/dL, confirming that eliminating feedback-driven error compounding is the dominant factor in reducing long-horizon bias. Third, including ODE features in the model provides consistent additional improvement in accuracy across all horizons, reducing ME by a further 1.90~mg/dL at 240 minutes and systematically bounding the error distribution in a way that is modest in isolation but consistent across all 48 horizons. Figure~\ref{fig:mae_me} shows these metrics across all horizons.

We investigated whether the ODE features helped to reduce outlier error by exploring accuracy performance on the top 100 worst-performing shared segments. Outlier error primarily occurred when unanticipated meals or other events occurred during the forecasting window. In these cases, all models performed poorly on the forecasting. On these segments, PhysioSeq2Seq had a lower RMSE of 183.8~mg/dL compared to 203.5~mg/dL for the recursive LSTM. This indicates that including the ODE states helped to limit large errors by providing a physiological upper bound that prevented large prediction errors caused by unanticipated meal events in the forecasting window (Appendix~\ref{app:worst_case}). An ablation study confirmed that the Seq2Seq architecture accounts for the dominant gain (ME improves from $-$24.51 to $-$12.52~mg/dL at 240 minutes), while use of ODE features provided consistent additional reductions in bias and error spread across all 48 horizons (Appendix~\ref{app:ablation}).


\section{Discussion}
\label{sec:discussion}
We show that a seq-to-seq architecture combined with features from an ODE model yields strong forecasting performance. The most significant finding is the 11.99~mg/dL reduction in 240-minute ME achieved by replacing the recursive LSTM with a Seq2Seq encoder--decoder, without any physiological augmentation. In the recursive strategy, each predicted $\hat{g}_{t+i}$ is fed back as input to the next step, so a small per-step downward bias compounds to $-$24.51~mg/dL after 48 iterations. The Seq2Seq decoder avoids feedback, making each output a function of the fixed encoder context and future covariates, with no path through previously generated predictions. ODE feature injection further improves performance by utilizing physiological models not available to data-driven models. The remote insulin action states $X_1, X_2, X_3$ and gut absorption states $M_1, M_2$ are dominant predictors of glucose trajectory over 1--4 hours, and the smooth future ODE trajectory acts as an implicit regularizer that discourages erratic high-variance predictions. Although $S_1$ and $S_2$ encode subcutaneous insulin kinetics, IOB provides complementary information through its 60-minute reference lag and patient-normalized scale. An ablation removing IOB while retaining all 10 ODE states degraded performance, confirming that the two representations are not interchangeable (Appendix~\ref{app:ablation}).

This work has three main limitations that motivate future extensions. First, future ODE digital twin states are propagated under basal-only conditions, so the twin cannot anticipate upcoming meals or correction boluses, producing a conservative baseline rather than systematic bias. Integrating probabilistic meal forecasts~\citep{Clara2023} as future covariates is a natural extension that could further reduce post-prandial error. Second, twin matching relies on greedy RMSE minimization over a fixed population of 300 ODE digital twin parameterizations. Despite this, the model generalizes well across 74 held-out patients. Incorporating patient metadata at matching time is an important future development that would utilize the same Seq2Seq architecture. Third, the virtual population of 300 digital twins varies primarily across insulin sensitivity parameters without systematic coverage criteria, and twin initialization applies a steady-state approximation even when glucose is actively rising or falling. This narrow physiological diversity and crude initialization may artificially depress standalone ODE performance, and future work should explore broader parameter sampling strategies and dynamic initialization methods.


\section{Conclusion}
\label{sec:conclusion}
We introduced PhysioSeq2Seq, a hybrid architecture that couples a population-matched ODE digital twin with a Seq2Seq LSTM. Our results show that architectural feedback, rather than model capacity, dominates long-horizon error, and that ODE digital twins, even under constrained population diversity and steady-state initialization assumptions, are effective feature extractors in hybrid forecasting pipelines. While incorrect forecasts in such systems could theoretically lead to suboptimal insulin dosing decisions, PhysioSeq2Seq is designed for use in decision-support that will require thorough clinical validation before real-world deployment, with direct implications for safer automated insulin delivery and hypoglycemia prevention in T1D. The framework is a promising candidate for domains where interpretable mechanistic models exist to augment data-driven forecasting.


\bibliographystyle{plainnat}
\bibliography{references}

\newpage

\appendix

\section{Twin Matching Algorithm}
\label{app:algorithm}
\begin{algorithm}[h]
  \caption{Twin matching and ODE state extraction.}
  \label{alg:twin_matching}
  \begin{algorithmic}[1]
    \REQUIRE Decision time $t$, CGM series $g^{\text{obs}}$,
    insulin events $\mathcal{E}$,
    active insulin time $\tau_{\text{act}}$,
    twin population $\{\boldsymbol{\theta}_j\}_{j=1}^{300}$

    \STATE \textbf{// Pass 1: Twin Matching}
    \STATE Resample CGM and insulin to 5-min grid over $[t_0, t]$
    \STATE Compute glucose slope $\dot{g}_0$ via linear regression
    over $[t_0,\; t_0 + 1\;\text{hr}]$
    \FOR{$j = 1$ \TO $300$}
    \STATE Initialize state $\mathbf{x}_{t_0}^{(j)}$ via
    $\textsc{SteadyState}(g_{t_0},\; \dot{g}_0,\; u_{\text{basal},t_0})$
    \STATE Simulate ODE$(\boldsymbol{\theta}_{j},\; \tau_{\text{act}})$
    over $[t_0, t]$
    \STATE $\epsilon_j \leftarrow \text{RMSE}(\hat{g}^{(j)},\; g^{\text{obs}})$
    \ENDFOR
    \STATE $j^* \leftarrow \arg\min_j\; \epsilon_j$

    \STATE \textbf{// Pass 2: History State Extraction}
    \STATE Re-initialize twin $j^*$ at $t_0$ via
    $\textsc{SteadyState}(g_{t_0},\; \dot{g}_0,\; u_{\text{basal},t_0})$
    \STATE Simulate twin $j^*$ over $[t_0, t]$ with full state
    logging at every timestep
    \STATE Extract $\mathbf{X}^{\text{hist}} \leftarrow
      \{\mathbf{x}_k\}_{k=t_0}^{t} \in \mathbb{R}^{L \times 10}$

    \STATE \textbf{// Pass 3: Forecast State Extraction}
    \STATE Compute $\bar{u}_{\text{basal}} \leftarrow
      \frac{1}{|\mathcal{T}_h|}\sum_{k \in \mathcal{T}_h} u_{\text{basal},k}$
    \STATE Compute $\dot{g}_t$ via linear regression over
    $[t - 1\;\text{hr},\; t]$ using observed CGM
    \STATE Compute $(\bar{Q}_1, \bar{Q}_2, \bar{S}_1, \bar{S}_2, \bar{I},
      \bar{X}_1, \bar{X}_2, \bar{X}_3)
      \leftarrow \textsc{SteadyState}(g_t,\; \dot{g}_t,\;
      u_{\text{basal},t})$
    \STATE Override subcutaneous chain from Pass~1 final state:
    $\bar{S}_1 \leftarrow S_1^{(\text{P1})},\;
      \bar{S}_2 \leftarrow S_2^{(\text{P1})}$
    \STATE Set $\mathbf{x}_t^{\text{init}} \leftarrow
      [\bar{Q}_1,\; \bar{Q}_2,\; \bar{S}_1,\; \bar{S}_2,\;
        \bar{I},\; \bar{X}_1,\; \bar{X}_2,\; \bar{X}_3,\;
        0,\; 0]^\top$
    \STATE Propagate twin $j^*$ for $H{=}48$ steps under constant
    $\bar{u}_{\text{basal}}$, zero bolus and carbohydrate inputs
    \STATE Extract $\mathbf{X}^{\text{fut}} \leftarrow
      \{\mathbf{x}_{t+i}\}_{i=1}^{H} \in \mathbb{R}^{H \times 10}$

    \ENSURE $\mathbf{X}^{\text{hist}}$, $\mathbf{X}^{\text{fut}}$
  \end{algorithmic}
\end{algorithm}
\section{Full ODE State-Space Matrix}
\label{app:ode_full}
Each virtual patient is characterized by 16 physiological parameters summarized in Table~\ref{tab:hovorka_params}.

\begin{table}[!htbp]
  \caption{
    ODE digital twin model parameters. Each virtual twin in the population of 300 is defined by a unique instantiation of these values.
  }
  \label{tab:hovorka_params}
  \centering
  \small
  \begin{tabular}{llll}
    \toprule
    \textbf{Symbol}  & \textbf{Description}                    & \textbf{Units}          \\
    \midrule
    $f_{c01}$        & Non-insulin-dependent glucose clearance & mmol/kg/min             \\
    $V_G$            & Glucose distribution volume             & L/kg                    \\
    $k_{12}$         & Transfer rate $Q_2 \to Q_1$             & min$^{-1}$              \\
    $a_G$            & Carb absorption rate constant           & dimensionless           \\
    $t_{\max,G}$     & Time of max gut absorption              & min                     \\
    $\mathrm{EGP}_0$ & Endogenous glucose production (basal)   & mmol/kg/min             \\
    $t_{\max,I}$     & Time of max insulin absorption          & min                     \\
    $k_e$            & Insulin elimination rate                & min$^{-1}$              \\
    $V_I$            & Insulin distribution volume             & L/kg                    \\
    $k_{a1}$         & Activation rate for $X_1$               & min$^{-1}$              \\
    $k_{a2}$         & Activation rate for $X_2$               & min$^{-1}$              \\
    $k_{a3}$         & Activation rate for $X_3$               & min$^{-1}$              \\
    $S_{F1}$         & Insulin sensitivity on glucose disposal & (mU/L)$^{-1}$min$^{-1}$ \\
    $S_{F2}$         & Insulin sensitivity on distribution     & (mU/L)$^{-1}$min$^{-1}$ \\
    $S_{F3}$         & Insulin sensitivity on EGP              & (mU/L)$^{-1}$min$^{-1}$ \\
    $\mathrm{BW}$    & Body weight                             & kg                      \\
    \bottomrule
  \end{tabular}
\end{table}
The continuous-time ODE digital twin system matrix $A_p(\xvec_k) \in \mathbb{R}^{10 \times 10}$ is:
\begin{equation}
  A_p =
  \resizebox{\textwidth}{!}{$
      \begin{pmatrix}
        -(X_1{+}k_{01}{+}k_r) & k_{12}          & 0                     & 0                        & 0
                              & -Q_1            & 0                     & -\mathrm{EGP}_0          & 0                     & K                     \\
        X_1                   & -(k_{12}{+}X_2) & 0                     & 0                        & 0
                              & Q_1             & -Q_2                  & 0                        & 0                     & 0                     \\
        0                     & 0               & -\frac{1}{t_{\max,I}} & 0                        & 0
                              & 0               & 0                     & 0                        & 0                     & 0                     \\
        0                     & 0               & \frac{1}{t_{\max,I}}  & -\frac{1}{t_{\max,I}}    & 0
                              & 0               & 0                     & 0                        & 0                     & 0                     \\
        0                     & 0               & 0                     & \frac{1}{t_{\max,I} V_I} & -k_e
                              & 0               & 0                     & 0                        & 0                     & 0                     \\
        0                     & 0               & 0                     & 0                        & S_{F1} k_{a1}
                              & -k_{a1}         & 0                     & 0                        & 0                     & 0                     \\
        0                     & 0               & 0                     & 0                        & S_{F2} k_{a2}
                              & 0               & -k_{a2}               & 0                        & 0                     & 0                     \\
        0                     & 0               & 0                     & 0                        & S_{F3} k_{a3}
                              & 0               & 0                     & -k_{a3}                  & 0                     & 0                     \\
        0                     & 0               & 0                     & 0                        & 0
                              & 0               & 0                     & 0                        & -\frac{1}{t_{\max,G}} & 0                     \\
        0                     & 0               & 0                     & 0                        & 0
                              & 0               & 0                     & 0                        & \frac{1}{t_{\max,G}}  & -\frac{1}{t_{\max,G}}
      \end{pmatrix}
    $}
\end{equation}
Here $k_r$ is a glucose renal clearance term active above $9$ mmol/L, and $K = a_G \cdot 0.18 \cdot t_{\max,G} \cdot \mathrm{BW}$ (mmol/kg) is the gut absorption scaling constant. The discrete-time matrices $A_d, B_d, D_d, G_d$ are obtained via:
\begin{equation}
  \exp\left(
  \begin{pmatrix}
      A_p & B_p & D_p(\xvec_k) & G_p \\
      \mathbf{0}_{3 \times 13}
    \end{pmatrix}
  \Ts
  \right)_{[0:10,\;:]},
\end{equation}
with $B_p = \mathbf{e}_3$, $G_p = \mathbf{e}_9$ (standard basis vectors), and $D_p(\xvec_k) = [X_1 Q_1 {-} k_d \mathrm{EGP}_0,{-}X_1 Q_1 {+} X_2 Q_2,\; 0, \ldots]^\top$.

\section{Dataset Quality Control Pipeline}
\label{app:qc}

Segments are excluded from all splits if any of the
following conditions are triggered:
\begin{enumerate}
  \item Fewer than 18 CGM points in the 3-hour history (minimum density requirement).
  \item CGM rate-of-change $> 40$ mg/dL per 5 minutes (sensor artifact or physical implausibility).
  \item Closest historical CGM $> 5$ minutes from decision time (stale sensor alignment).
  \item Residual NaN in CGM tail after linear interpolation over gaps of $\leq 3$ steps.
  \item Future CGM jump $> 40$ mg/dL between consecutive 5-minute observations.
  \item Twin matching failure (all 300 twins return $\epsilon_j = \infty$).
\end{enumerate}
These six criteria produce a 94.3\% extraction yield (289,859 valid of 307,474 scanned segments).

\section{Worst-Case Segment Analysis}
\label{app:worst_case}
To evaluate model robustness under challenging glycemic conditions, we analyze performance on the most difficult segments in the shared test set using two complementary strategies, summarized in Table~\ref{tab:worst_case}.

First, worst-case segments are identified independently for each model by ranking shared test segments according to per-segment RMSE and selecting the top $K = 100$ highest-error segments. Because this procedure is performed separately for each model, the resulting sets differ across models, characterizing model-specific failure modes rather than enabling direct comparison on identical inputs.

Second, to identify segments that are intrinsically difficult, we compute the average per-segment RMSE across all four models and select the top $K = 100$ segments with the highest average error. All models are evaluated on this identical set, enabling fair head-to-head comparison and isolating scenarios that are broadly challenging regardless of model architecture.

\begin{table}[H]
  \caption{
    Average RMSE, MAE, and ME over the top-100 worst-performing
    segments (independent per model, and shared across all models).
    All values are reported in mg/dL.
  }
  \label{tab:worst_case}
  \centering
  \small
  \begin{tabular}{lcccccc}
    \toprule
     & \multicolumn{3}{c}{\textbf{Independent Worst-100}}
     & \multicolumn{3}{c}{\textbf{Shared Worst-100}}                                           \\
    \cmidrule(lr){2-4}\cmidrule(lr){5-7}
    \multicolumn{1}{c}{\textbf{Model}}
     & RMSE                                               & MAE            & ME
     & RMSE                                               & MAE            & ME                \\
    \midrule
    Digital Twin
     & 255.4                                              & 235.4          & $-$194.2
     & 235.0                                              & 211.9          & $-$193.4          \\
    Recursive LSTM
     & 208.4                                              & 189.7          & $-$178.1
     & 203.5                                              & 185.6          & $-$170.5          \\
    Seq2Seq $-$ ODE
     & 192.3                                              & 175.4          & $-$175.1
     & 184.8                                              & 167.8          & $-$156.5          \\
    Seq2Seq $+$ ODE
     & \textbf{189.0}                                     & \textbf{171.1} & \textbf{$-$156.9}
     & \textbf{183.8}                                     & \textbf{166.0} & \textbf{$-$152.4} \\
    \bottomrule
  \end{tabular}
\end{table}

\section{Ablation Study}
\label{app:ablation}
All values are reported in mg/dL.
\begin{table}[H]
  \caption{Ablation results at the 240-minute forecast horizon.}
  \label{tab:ablation_240}
  \centering
  \small
  \begin{tabular}{lccc cccc}
    \toprule
    \multicolumn{1}{c}{\textbf{Model}}
     & \textbf{Seq2Seq} & \textbf{ODE}
     & \textbf{RMSE}    & \textbf{MAE}   & \textbf{ME}       & \textbf{IQR}   \\
    \midrule
    Digital Twin
     & $\times$         & $\checkmark$
     & 87.60            & 67.90          & $-$14.30          & 105.84         \\
    Recursive LSTM
     & $\times$         & $\times$
     & 61.51            & 44.11          & $-$24.51          & 65.83          \\
    Seq2Seq $-$ ODE
     & $\checkmark$     & $\times$
     & 54.84            & 39.54          & $-$12.52          & 61.47          \\
    Seq2Seq $+$ ODE
     & $\checkmark$     & $\checkmark$
     & \textbf{54.12}   & \textbf{39.28} & \textbf{$-$10.62} & \textbf{60.76} \\
    \bottomrule
  \end{tabular}
\end{table}

\begin{table}[h]
  \centering
  \caption{Ablation study evaluating the impact of ODE coupling and physiological input features across forecast horizons.}
  \label{tab:ablation_full}
  \small
  \setlength{\tabcolsep}{2pt}
  \begin{tabular}{lcccccccccccc}
    \toprule
     & \multicolumn{3}{c}{\textbf{30 min}}
     & \multicolumn{3}{c}{\textbf{60 min}}
     & \multicolumn{3}{c}{\textbf{120 min}}
     & \multicolumn{3}{c}{\textbf{240 min}}                                     \\
    \cmidrule(lr){2-4}\cmidrule(lr){5-7}\cmidrule(lr){8-10}\cmidrule(lr){11-13}
    \multicolumn{1}{c}{\textbf{Model}}
     & RMSE                                 & MAE            & ME
     & RMSE                                 & MAE            & ME
     & RMSE                                 & MAE            & ME
     & RMSE                                 & MAE            & ME               \\
    \midrule
    Seq2Seq $-$ IOB
     & 21.83                                & 15.16          & $-$2.89
     & 33.86                                & 24.05          & $-$4.99
     & 46.67                                & 33.94          & \textbf{$-$7.34}
     & 54.69                                & 40.17          & \textbf{$-$8.83} \\
    Seq2Seq $-$ ODE
     & 22.46                                & 15.71          & $-$3.62
     & 34.46                                & 24.64          & $-$6.22
     & 47.20                                & 34.00          & $-$10.08
     & 54.84                                & 39.54          & $-$12.52         \\
    Seq2Seq $+$ ODE
     & \textbf{21.71}                       & \textbf{15.12} & \textbf{$-$2.21}
     & \textbf{33.71}                       & \textbf{24.00} & \textbf{$-$4.73}
     & \textbf{46.16}                       & \textbf{33.40} & $-$8.04
     & \textbf{54.12}                       & \textbf{39.28} & $-$10.62         \\
    \bottomrule
  \end{tabular}
\end{table}

\begin{table}[h]
  \centering
  \caption{Error dispersion across forecast horizons: IQR and $[$min, max$]$ prediction error range.}
  \label{tab:ablation_dispersion}
  \small
  \setlength{\tabcolsep}{3pt}
  \begin{tabular}{lcccccccc}
    \toprule
                                       & \multicolumn{2}{c}{\textbf{30 min}}
                                       & \multicolumn{2}{c}{\textbf{60 min}}
                                       & \multicolumn{2}{c}{\textbf{120 min}}
                                       & \multicolumn{2}{c}{\textbf{240 min}}                     \\
    \cmidrule(lr){2-3}\cmidrule(lr){4-5}
    \cmidrule(lr){6-7}\cmidrule(lr){8-9}
    \multicolumn{1}{c}{\textbf{Model}} & IQR                                  & {[Min, Max]}
                                       & IQR                                  & {[Min, Max]}
                                       & IQR                                  & {[Min, Max]}
                                       & IQR                                  & {[Min, Max]}      \\
    \midrule
    Digital Twin
                                       & 26.13                                & [$-$221.6, 198.4]
                                       & 50.35                                & [$-$272.6, 314.2]
                                       & 83.12                                & [$-$380.3, 328.8]
                                       & 105.84                               & [$-$398.3, 367.5] \\
    Recursive LSTM
                                       & 23.66                                & [$-$236.5, 188.5]
                                       & 37.72                                & [$-$244.0, 284.2]
                                       & 53.89                                & [$-$306.1, 289.9]
                                       & 65.83                                & [$-$303.7, 216.1] \\
    Seq2Seq $-$ IOB
                                       & 20.59                                & [$-$208.4, 179.6]
                                       & 34.41                                & [$-$238.6, 282.3]
                                       & 51.82                                & [$-$278.2, 281.5]
                                       & 62.95                                & [$-$278.0, 187.2] \\
    Seq2Seq $-$ ODE
                                       & 22.07                                & [$-$217.2, 154.3]
                                       & 35.94                                & [$-$235.7, 254.5]
                                       & 51.60                                & [$-$279.7, 253.9]
                                       & 61.47                                & [$-$290.1, 221.2] \\
    Seq2Seq $+$ ODE
                                       & \textbf{20.73}                       & [$-$200.4, 179.8]
                                       & \textbf{34.30}                       & [$-$236.3, 286.8]
                                       & \textbf{50.30}                       & [$-$279.6, 281.2]
                                       & \textbf{60.76}                       & [$-$289.6, 227.0] \\
    \bottomrule
  \end{tabular}
\end{table}

\section{Qualitative Forecast Trajectories}
\label{app:trajectories}
\begin{figure}[h]
  \centering
  \includegraphics[width=\linewidth]{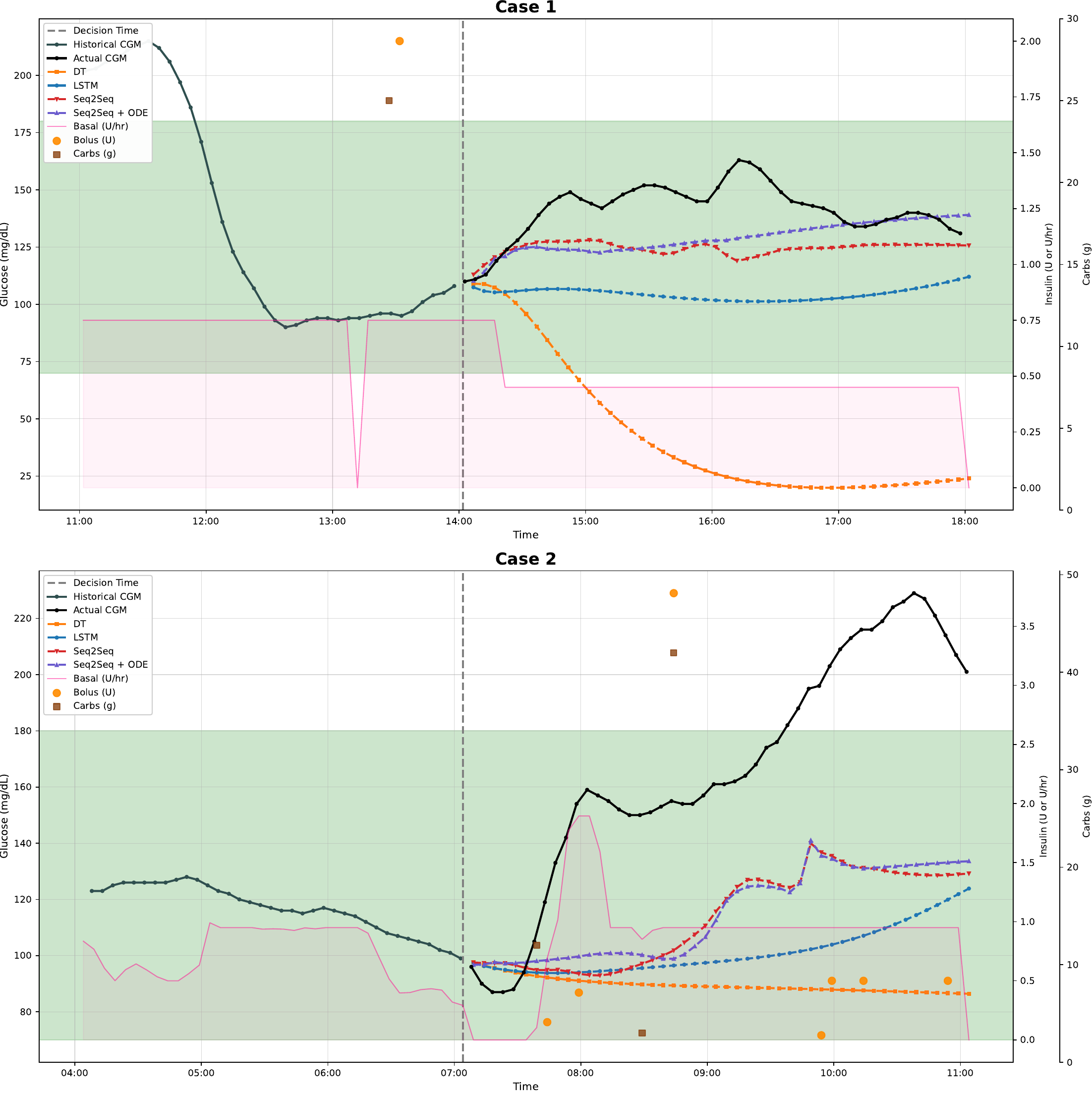}
  \caption{
    Representative forecast trajectories from two test-set segments. Case~1: A well-conditioned segment where Seq2Seq $+$ ODE most closely tracks the actual CGM response, while the recursive LSTM exhibits a systematic negative bias. Case~2: A segment with an unexplained post-decision disturbance and no logged meal at $t_0$, where all models fail to anticipate the glucose excursion. Shaded regions indicate the target glucose range (70--180~mg/dL). The dashed vertical line marks the decision time $t_0$.
  }
  \label{fig:case_trajectories}
\end{figure}

Figure~\ref{fig:case_trajectories} illustrates representative forecast trajectories from two segments in the held-out test set, demonstrating how each model behaves under contrasting conditions.

Case 1 shows a segment where all inputs are well-observed. The Seq2Seq $+$ ODE model most closely tracks the actual CGM trajectory, capturing both the timing and magnitude of the post-meal glucose rise. The recursive LSTM exhibits a systematic negative bias, consistently underestimating glucose across the forecast horizon. The Digital Twin produces a physics-driven trajectory that diverges from the observed response.

Case 2 illustrates a segment where an unexplained disturbance occurs after the decision time. No meal event was recorded in the structured input at $t_0$, leaving all models without the information needed to anticipate the excursion. As a result, every model fails to predict the sharp glucose rise, and the actual CGM climbs well above all forecasts.

\end{document}